\documentclass{article}

\usepackage{arxiv}

\usepackage[utf8]{inputenc} 
\usepackage[T1]{fontenc}    
\usepackage{hyperref}       
\usepackage{url}            
\usepackage{booktabs}       
\usepackage{amsfonts}       
\usepackage{nicefrac}       
\usepackage{microtype}      
\usepackage{lipsum}
\usepackage{graphicx}
\graphicspath{ {./images/} }
\usepackage{float}
\usepackage{amsmath}
\usepackage{subfig}
\usepackage{multirow}
\usepackage{color}
\usepackage{lscape,adjustbox}
\usepackage{float}
\usepackage{footnote}
\usepackage{amssymb}
\usepackage{threeparttable}
\usepackage{enumitem} 
\usepackage{tabu}
\usepackage{hhline,colortbl}
\usepackage{array}
\usepackage{algorithm}

\title{Pain Analysis using Adaptive Hierarchical Spatiotemporal
Dynamic Imaging}

\author{
   Issam Serraoui \\
  IEMN - DOAE UMR CNRS 8520- \\, Université Polytechnique Hauts-de-France, \\Valenciennes, France.\\ Laboratoire d'imagerie,\\ de vision et d'intelligence artificielle,\\ École de technologie supérieure,\\ Université du Québec Montreal, Canada\\
  \texttt{issam.serraoui@uphf.fr} \\
 \\
   \And
   Eric Granger \\
  Laboratoire d'imagerie,\\ de vision et d'intelligence artificielle,\\ École de technologie supérieure,\\ Université du Québec Montreal, Canada\\
  \texttt{Eric.Granger@etsmtl.ca} \\
   \And
Abdenour Hadid \\
  Sorbonne Center for Artificial Intelligence\\
    Abu Dhabi, UAE.\\
  \texttt{abdenour.hadid@ieee.org} \\
   \And
   Abdelmalik Taleb-Ahmed \\
  IEMN - DOAE UMR CNRS 8520- \\, Université Polytechnique Hauts-de-France, \\Valenciennes, France.\\
  \texttt{abdelmalik.taleb-ahmed@uphf.fr} \\
}

\begin{document}
\maketitle
\begin{abstract}
Automatic pain intensity estimation plays a pivotal role in healthcare and medical fields. While many methods have been developed to gauge human pain using behavioral or physiological indicators, facial expressions have emerged as a prominent tool for this purpose. Nevertheless, the dependence on labeled data for these techniques often renders them expensive and time-consuming. To tackle this, we introduce the Adaptive Hierarchical Spatio-temporal Dynamic Image (AHDI) technique. AHDI encodes spatiotemporal changes in facial videos into a singular RGB image, permitting the application of simpler 2D deep models for video representation. Within this framework, we employ a residual network to derive generalized facial representations. These representations are optimized for two tasks: estimating pain intensity and differentiating between genuine and simulated pain expressions. For the former, a regression model is trained using the extracted representations, while for the latter, a binary classifier identifies genuine versus feigned pain displays. Testing our method on two widely-used pain datasets, we observed encouraging results for both tasks. On the UNBC database, we achieved an MSE of 0.27 outperforming the SOTA which had an MSE of 0.40. On the BioVid dataset, our model achieved an accuracy of 89.76\%, which is an improvement of 5.37\% over the SOTA accuracy. Most notably, for distinguishing genuine from simulated pain, our accuracy stands at 94.03\%, marking a substantial improvement of 8.98\%. Our methodology not only minimizes the need for extensive labeled data but also augments the precision of pain evaluations, facilitating superior pain management.
\end{abstract}


\section{INTRODUCTION}
{Pain, an essential indicator of potential bodily issues, plays a significant role in medical diagnostics and treatment \cite{Lynch200185}. The ability to automatically evaluate pain, especially through facial expressions, offers vital insights into an individual is well-being \cite{smith2019understanding, ekman2002facial}. This evaluation primarily hinges on identifying Facial Action Units (AUs), specific facial muscle movements tied to expressions of pain \cite{prkachinXXXXstructure}. Though traditional methods leveraged labeled datasets for this, they often proved resource-intensive. Consequently, there has been a shift towards deep learning and video representation techniques to enhance pain detection efficacy.}

{In the realm of pain evaluation, historical methodologies primarily focused on extracting spatiotemporal features from individual video frames or treating videos as volumetric entities \cite{lucey2012painful,kaltwang2012continuous}. With the evolution of deep learning, the field has witnessed revolutionary advancements, notably through the application of Convolutional Neural Networks (CNNs) and Recurrent Neural Networks (RNNs). Such advancements have led to state-of-the-art techniques employing Recurrent Convolutional Neural Networks (RCNNs) and hybrids of CNNs with Long Short-Term Memory (LSTM) networks, reshaping facial video pain analysis \cite{bromley1994signature,zhou2016recurrent,rodriguez2017deep}.}

A persistent challenge, however, is the labor-intensive task of data annotation via the Facial Action Coding System (FACS), which demands specific expertise and time. 
\begin{figure}[!t]
   \centering
   \includegraphics[width=0.5\textwidth]{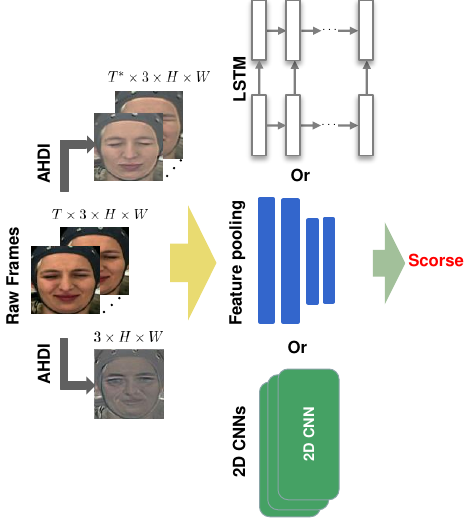}
\caption{{Schematic representation of the AHDI transformation process: A dual-pathway approach where raw video data is encoded into either multiple dynamic images for sequential analysis or a singular dynamic image for spatial analysis} }

   \label{fig0}
\end{figure}
{This rigorous methodology segments facial expressions into action units, each described by onset, peak, and intensity \cite{ekman1978facial,littlewort2009automatic}. The inherent limitations of supervised learning techniques, particularly their dependence on training data, further complicate the adaptation to diverse datasets, spotlighting an urgent need for richer datasets and more versatile evaluation techniques \cite{othman2019cross,wang2018unsupervised,yan2016transfer}. Recognizing the potential of self-supervised learning, recent trends are exploring its application as an effective substitute to conventional supervised methods for automated pain assessment, thus mitigating the reliance on extensive labeled data \cite{jing2020selfsupervised,tavakolian2020self}.}

{With an increasing dependency on videos, it is crucial to achieve a comprehensive understanding of them. While advancements in deep CNNs have significantly improved image understanding \cite{krizhevsky2012imagenet}, video comprehension offers myriad perspectives—ranging from viewing videos as sequences of images to understanding them as attribute subspaces \cite{karpathy2014large,jain2014better}. Yet, determining the most efficient video representation remains a formidable research challenge.}

Building upon the foundational work of \cite{bilen2016dynamic}, we extended their insights to pain estimation through the deployment of the Adaptive Hierarchical Spatio-temporal Dynamic Image (AHDI) methodology. The AHDI model effectively captures spatiotemporal nuances of facial videos in a unified RGB frame, {thereby streamlining the application of 2D deep learning architectures. Figure \ref{fig0} offers a visualization of our proposed AHDI process. From an input video, AHDI can create either multiple dynamic images suited for LSTM processing or a singular dynamic image apt for 2D CNN interpretations. This adaptability emphasizes AHDI is robustness in different deep learning contexts, making it versatile and potent. Our unique approach views video sequences as dynamic images, providing a comprehensive view of both appearance and motion. Harnessing the power of AHDI, subtle facial expression variations indicative of pain can be meticulously recorded. When coupled with our specialized residual network, the dynamic images amplify finer facial movements, enhancing their distinguishing capabilities.}

{Grounded in labeled data from two prominent pain datasets, our training methodology encompasses both a regression model for pain assessment and a binary classifier to distinguish genuine from feigned pain expressions. Preliminary evaluations reveal promising results. The synergistic application of AHDI with residual networks not only alleviates data labeling burdens but also augments the accuracy of pain assessments.}

In summary, the salient contributions of this paper are:
\begin{itemize}
\item The inception of the {Adaptive Hierarchical Spatio-temporal Dynamic Image (AHDI), an innovative RGB representation capturing an entire video is intricacies.} We aspire to integrate video dynamics directly at the image level.
\item The synthesis of these observations to formulate an advanced neural framework optimized for {end-to-end video training. Melding frame-based static data with video dynamics, we have set new performance standards, outclassing prevailing benchmarks.}
\item An exhaustive exploration of vital design {elements, notably temporal window spans, sampling rates, and temporal pooling techniques.}
  \item Comparison against state-of-the-art approaches showcased the versatility and effectiveness of our proposed model. With our novel methodologies, we have been able to achieve superior performance in differentiating posed pain expressions from genuine ones.
\item  The presentation of extensive experimental results that underscore the advantage of AHDI, especially when coupled with advanced neural architectures. The study provides empirical evidence of the method is potential, paving the way for future research in the domain.
\end{itemize}

\section{Related work}

{Pain assessment research primarily falls into three categories}: behavioral-based, physiological-based, and a multimodal approach. However, assessing pain via facial imagery is intricate. {Among the key challenges are:}
\begin{itemize}

    \item     {Inconsistencies in facial image quality due to variations in lighting, angles, and backgrounds.}

    \item  {The effect of external factors, like smiling during pain or gender-related pain expression differences}, pose challenges in accurately determining pain levels solely from facial expressions.

   \item {An ambiguous correlation between facial cues and pain rating scales remains ambiguous \cite{prkachinXXXXstructure}. Evidence indicates varied pain expressions between genders for the same stimulus \cite{wang2006hidden}, and facial expressions often don't align closely with self-reported pain levels \cite{prkachinXXXXstructure}.}
\end{itemize}
An effective method for autonomously gauging self-reported pain from facial cues {should account for individual expressiveness nuances and address challenges like image inconsistencies, external influences, and the vague link between facial cues and pain scales.}

{While there is not a universally accepted model for emotion intensity, pediatric machine learning models have made strides in interpreting pain responses through facial expressions, auditory cues, and physiological metrics like GSR (Galvanic Skin Response). By integrating these indicators in machine learning, particularly in automated pain detection systems, we approach a refined methodology for pain intensity estimation~\cite{martinez2017personalized,thiam2017multi,tavakolian2019spatiotemporal,tavakolian2020selfsupervised,xin2020pain}. These data-driven systems have transformative potential in pain diagnosis and management.}

To gain a holistic understanding of the pain estimation paradigm, one can dissect the literature, with an emphasis on various phases, particularly accentuating feature extraction and pre-processing methodologies:
{Historically, studies used motion-centric techniques, primarily optical flow, to determine pixel velocities across video sequences, ensuring a consistent pixel-wise relationship \cite{lopez2017personalized, simonyan2014two}.}

Recent advancements in pain analysis emphasize the utility of innovative models and techniques. {Lucey {\it et al.} demonstrated the effectiveness of Active Appearance Models (AAMs), emphasizing the importance of canonical normalized appearance (CAPP) features in pain recognition \cite{lucey2011painful}. Furthermore, energy-centric representations like the Gabor amplitude outperformed grayscale representations in facial action recognition \cite{lucey2010registration}. As an enhancement, Hammal {\it et al.} integrated results from Log-Normal filters, refining the automatic recognition of pain intensity \cite{hammal2012automatic}.}

The modern era has witnessed a surge in endeavors to encapsulate temporal information from video sequences through the lens of deep neural networks. Illustratively, {Zhou {\it et al.} \cite{zhou2016recurrent} proposed a 2D Recurrent Convolutional Neural Network (RCNN) for pain estimation, which treats video frames as vectors, possibly compromising facial data is structural integrity. Meanwhile, Rodriguez {\it et al.} combined 2D CNN features with LSTM models to gain temporal insights from video sequences \cite{rodriguez2017deep,hochreiter1997long}. Venturing deeper, Haochen {\it et al.} introduced a deep attention transformer-based network, distinguishing spatial and temporal feature extraction for pain estimation \cite{xu2021deep}. The design effectively captures relationships across extended video sequences, prioritizing significant frames.}

Image analysis has undergone significant transformation with 3D Convolutional Neural Networks (3D-CNNs) emerging as alternatives to the conventional 2D counterparts. Zhao {\it et al.} \cite{zhao2018learning} and Rajasekhar {\it et al.} \cite{rajasekhar2021deep} introduced a 3D-CNN architecture that integrates static and dynamic facial features, and further uses optical flow sequences to highlight facial expressions.

{Wang {\it et al.}} \cite{wang2018pain} {developed a 3D-CNN model combining deep spatiotemporal and handcrafted features, enhancing pain intensity estimation from video sequences. This model uniquely captures appearance and motion, emphasizing its role in video analysis, especially action recognition.}

Gnana {\it et al.} \cite{rajasekhar2021deep} proposed the Weakly-Supervised Domain Adaptation with Ordinal Regression (WSDA-OR) model. {Designed for videos with sparse coarse labels, this model accentuates ordinal relationships between frames and employs soft Gaussian labels for efficient sequence-level labeling.}

{Lastly, Tavakolian {\it et al.} \cite{tavakolian2019spatiotemporal} presented a 3D model, optimized for depicting dynamic spatiotemporal facial variations in videos. Their novel approach involves training 3D convolutional networks using pre-existing 2D architectures, demonstrating a practical method for 3D training and underscoring the benefits of transfer learning for comprehensive video analysis.}
3D-CNN research endeavors to enhance traditional CNNs by adding a temporal dimension, transitioning from 2D to 3D filters \cite{wang2018pain}. This shift hasn't drastically reaped dividends, possibly due to the dearth of labeled video content. Enhancing annotated video content might address this \cite{Tran_2015_ICCV}, yet ensuring spatial consistency in frames seems crucial. {Pixel-to-pixel registration via a video is optical flow} and the integration of action tubes have proven fruitful in capturing pain-specific facial Action Units (AUs) \cite{tavakolian2019learning,wang2018pain}.

Historical roots of action recognition in CNNs stem from architectures utilizing dynamic images. Vedaldi {\it et al.} \cite{vedaldi2016dynamic} posited that short frame sequences aptly capture temporal action structures. The concept of rank pooling, central to their dynamic images, had its origins in hand-crafted frame representation \cite{fernando2015modeling,fernando2016rank}. Later studies evolved the capabilities of rank pooling via hierarchical approaches \cite{fernando2016discriminative}.

In this context, we present the {Adaptive Hierarchical Spatio-temporal Dynamic Image (AHDI), a novel method for video representation. AHDI condenses spatiotemporal variations from facial videos into a single RGB image, facilitating simpler 2D models for video representation. This technique has multiple benefits:}

1.    The resultant RGB image, akin to still image designs, can be smoothly processed by CNNs, enabling extraction of video-specific dynamic features.

2.    {AHDI is efficiency is evident. Deriving the dynamic image is fast and simple, transforming video classification 
\begin{figure}[H]
   \centering
   \includegraphics[width=0.4\textwidth]{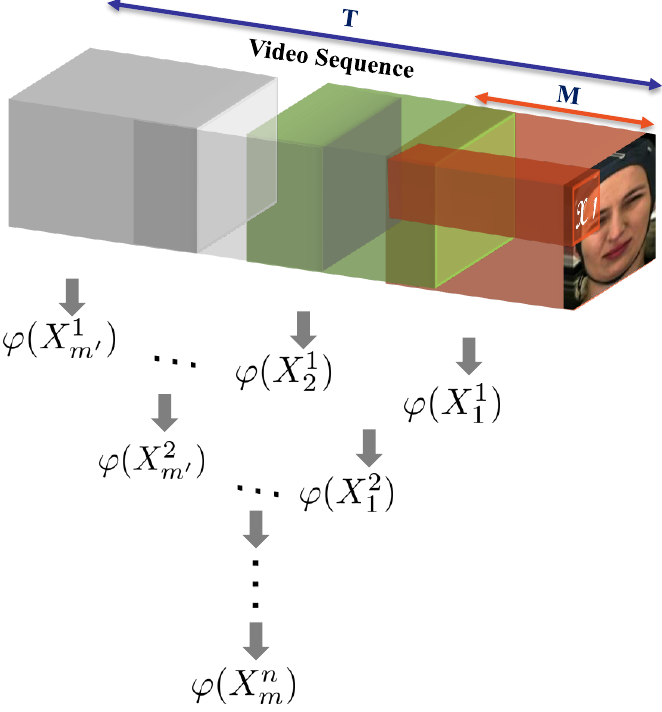}
\caption{Illustration of the AHDI approach: Overlapping video segments of variable lengths are compared sequentially to generate an image that captures both appearance and dynamic transitions between segments.}
   \label{fig2}
\end{figure}
to a single-image classification using a standard CNN.}

3.    A notable compression aspect of this method is encapsulating an entire video within the data of a single frame.

{With these novel advancements in feature extraction and pre-processing, the aim is to develop comprehensive pain assessment models. Delving into these groundbreaking studies provides insights into the potential of various pain estimation techniques from facial imagery, refining pain treatment strategies.}
\section{Proposed approach}
{We present a comprehensive framework pivoted on the notion of a "dynamic image," an RGB snapshot that merges both aesthetic and dynamic elements of a video sequence. 
This is especially salient for capturing fleeting facial changes indicative of pain, as schematized in Fig. \ref{fig2}.}

{A continuous frame sequence from the video undergoes the AHDI transformation, amplifying subtle facial dynamics. These dynamic images subsequently serve as the cornerstone for our residual neural architecture. The network aims at two primary goals: (1) to evaluate the intensity of pain, and (2) to differentiate between genuine and staged expressions of pain.}

Although CNNs are esteemed for their capacity to autonomously generate potent data representations, they are often bounded by their preset architectures. A pivotal challenge in adapting a CNN for video data is optimally feeding video information to it. Previous studies, as cited in \cite{fernando2016discriminative,fernando2015modeling}, typically suggested encoding video segments as arrays or utilizing recurrent designs. We propose an innovative departure: condensing the video narrative into one image, suitable for processing by simpler, yet effective, CNN structures like ResNets \cite{he2016deep}. {This approach not only boosts efficiency but also aligns seamlessly with conventional neural frameworks.}
\begin{figure*}[!t]
   \centering
   \includegraphics[width=0.75\textwidth]{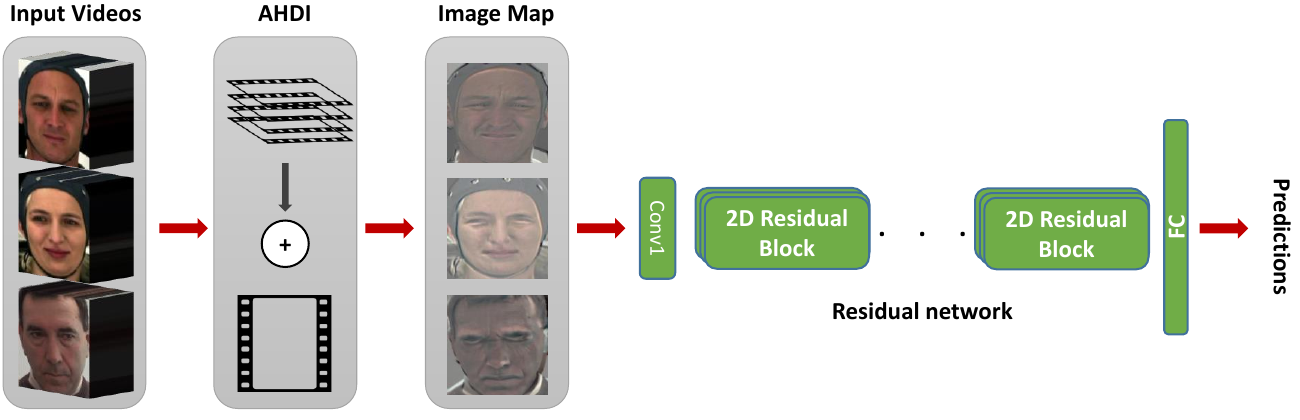}
   \caption{General structure of Proposed framework for pain estimation.}
   \label{fig3}
\end{figure*}
\subsection{Adaptive Hierarchical Spatio-temporal Dynamic Image}
\label{sec_AHDI}
In alignment with the methodology in \cite{bilen2016dynamic}, constructing a unified static image to represent the dynamic nature of a video poses intricate challenges. A primary complication arises from the way video frames embed visual data. These frames
can overshadow the dynamics crucial for action recognition.
Consider a set of training videos, denoted as 
\[
\mathcal{Y}=\{ Y_1,Y_2,\ldots ,Y_i \}
\]
Here, each video is classified into one of \(h\) classes. Each video, represented by \(Y_j\), consists of a sequence of frames 
of length \(T\), so we can express:
\[
{Y_j}=\{ I_1, I_2,\ldots, I_T \}
\] and  its sub-sequences 
\[{Y_j}= \left\{X_{1},X_{2},\cdots ,X_{m}  \right\} \]
The frames follow a strict chronological order, as \( I_1\prec I_2\ldots \prec I_T\). 

For encoding, we initiate by computing the information content weight. This is achieved by navigating a sliding window across each set of \(M\) consecutive frames, wherein the window spans an \(H \times W \times D\) spatial-temporal neighborhood at each juncture (refer Fig. \ref{fig2}-\ref{fig3}). The culmination of this process is an information weight map for every pair of overlapping video segments.
Consider \({(x_{(i,j)})}^{p}_{q}\) as the \((i,j)\)-th points within the reference frame sequence \(X^p_q\), where \(p \in \{0,n\}\) and \(q \in \{0,m\}\). Here, \(n\) symbolizes the number of hierarchical layers and \(m\) represents the segments count.
Mapping between \(M\) consecutive frames in overlapping segments is given by:
\begin{equation} \label{eq5}
\hat{\rho}(i_1,i_2,\ldots,i_M; \varphi)^{p}_{q} = \sum^{M}_{\tau=1} \alpha_\tau\varphi(i_\tau)
\end{equation}
where the coefficients $\alpha_\tau$ are given by:
\begin{equation} \label{eq3}  
\alpha_\tau=2(M  - \tau + 1) - (M + 1)~(H_M- H_{\tau-1}),
\end{equation}\\
where $H_\tau = \sum^{\tau}_{i=1}  1/\tau$ is the $ \tau$ -th Harmonic number and
$H_0=0$ and $M$ is number of frames.\\
Consequently, the derivation for \( \varphi(X^{p}_{q})\) is as:
\begin{equation} \label{eq6}
\varphi(X^{p}_{q}) =
\begin{bmatrix}
\varphi{(x_{(1,1)})}^{p}_{q} & \ldots & \varphi{(x_{(1,o)})}^{p}_{q}\\
\vdots & \ddots & \vdots\\
\varphi{(x_{(o,1)})}^{p}_{q} & \ldots & \varphi{(x_{(o,o)})}^{p}_{q}
\end{bmatrix}, \quad i,j \in \{1, \ldots, o\}
\end{equation}
with \(o\) being the patches count.

Amalgamating the above formulations and incorporating Algorithm 1 is results, the final mutual information weight function (AHDI) can be streamlined to:
\begin{equation} \label{eq7}
AHDI = \sum^{m}_{q=1,p=1} \alpha_{q,p}\varphi(\varphi(X^{p}_{q}))
\end{equation}
Here, \(m\) denotes the video segments count.
On completion of this for all consecutive segments, an \((N-1)\)-channel image map is generated, primed for facial expression analysis. 
To reiterate, encoding appearance and dynamics from videos necessitates a strategic capture of spatiotemporal intricacies. We adopt a strategy rooted in the principle that descriptive regions are visually more salient \cite{khatoonabadi2015bits}. Building upon information theory, an image is local content can be assessed in terms
of bit sequences.
Our methodology segments video frames contingent on in-scene motion activity (refer to Algorithm \ref{alg:AHDI}). The segment is length is flexible, adapting to the scene is motion intensity. For instances with heightened motion, shorter segments encapsulate the swift actions. Adjustments to the segment are based on scene intricacy or the granularity needed for subsequent processing. A static, minimalist scene might utilize a lengthier segment compared to a detail-rich, dynamic one.
\begin{center}
\scalebox{0.8}{
     \begin{minipage}{1\linewidth}
\begin{algorithm}[H]
    \caption{AHDI}
    \label{alg:AHDI}

    \textbf{Input:} 
    \begin{itemize}
        \item Video \(Y\) of \(T\) frames
        \item Threshold value \(\epsilon\)
        \item Primary number of segments \(n\)
    \end{itemize}

    \textbf{Output:} 
    \begin{itemize}\item Segmented sequences
\item Final AHDI image\end{itemize}
    \textbf{Procedure:}
    \begin{enumerate}
        \item \textbf{Compute Motion Activity:}
        \begin{enumerate}
            \item For each frame, compute optical flow \(OP(u, v)\) between \(I_{t-1}(x, y)\) and \(I_t(x, y)\) using Farneback is method.
            \item Convert \(OP(u, v)\) to magnitude \(m\) and angle \(\theta\).
            \item Calculate motion activity \(MA_t\) for frame \(t\): \(MA_t = \sum_{x,y} m(x, y)\).
            \item If \(MA_t > \epsilon\), frame is active; else, it is inactive.
        \end{enumerate}
        
        \item \textbf{Segmentation Based on Motion:}
        \begin{enumerate}
            \item Cluster frames into \(n\) segments based on motion activity.
            \item Calculate segment mean \(SM_i\): \(SM_i = \frac{\sum_t MA_t \times [SL_t == i]}{|{t: SL_t == i}|}\).
            \item Calculate average motion across all segments: \(ASM = \frac{1}{n} \sum_i SM_i\).
            \item For each segment, determine segment length \(L_i\): \(L_i = \frac{SM_i}{ASM} \times T\).
            \item For each segment, compute starting and ending indices: \(start\_idx, end\_idx\).
        \end{enumerate}
        
        \item Compute motion profile \( \varphi{(x_{(i,j)})}^{p}_{q} \) using Eq. \ref{eq5}.
        \item Iterate over all segments to refine using Eq. \ref{eq7}.
    \end{enumerate}
\end{algorithm}

\end{minipage}%
}
\end{center}
{\section{Unified and Segmented Dynamic Images Architectures}}
{Dynamic visualizations, encompassing both images and maps, can be synthesized from video sequences of diverse durations. Several neural network architectures are proposed herein, capable of assimilating single or multiple dynamic visual elements across varying video durations.}

{\subsection{Single Dynamic Visualization (AHDI)}}

{Within this methodology, a solitary dynamic image or map epitomizes the holistic content of a video. Deploying a Convolutional Neural Network (CNN) on these dynamic images facilitates extraction of temporal patterns inherent to the video content. An innate advantage is the potential to initialize with a CNN pre-trained on static imagery, e.g., ResNet on ImageNet data. Ensuing fine-tuning on dynamic imagery enables the CNN to capture video-centric dynamics, alleviating the need for exhaustive retraining often hindered by data paucity.}
\begin{figure*}[!t]
   \centering
   \includegraphics[width=0.8\textwidth]{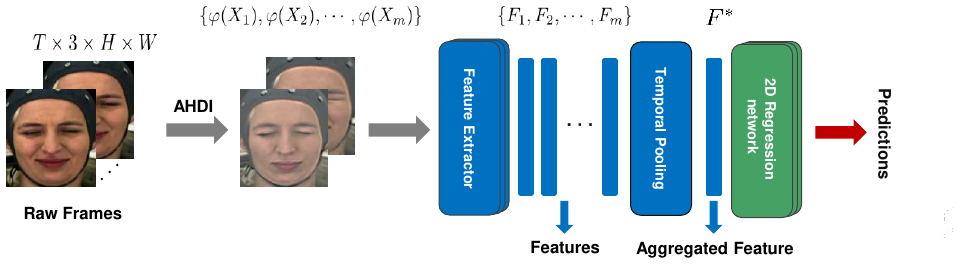}
   \caption{Illustration of Multiple Dynamic Visualizations architectures for pain estimation.}
   \label{fig4}
\end{figure*}
{\subsection{Multiple Dynamic Visualizations (MDV)}}

{Despite the data efficiency of fine-tuning, the disparity between conventional and dynamic imagery implies a requisite for substantial annotated datasets, a feat challenged by limited video data accessibility. }

{To address this, we advocate for video segmentation to generate multiple dynamic images or maps. Herein, each video undergoes segmentation into overlapping sectors defined by duration \( M \) and stride \( s \). This strategy effectively augments the dataset by a factor of approximately \( \frac{T}{M} \), where \( T \) is the average frame count per video. It resembles data augmentation but leverages frame subsets instead of common image modifications.
In other words, we can stop at the first hierarchical layer of AHDI (See Fig. \ref{fig2})}. {The final mutual information weight function Multiple Dynamic Visualizations (MDV) can be streamlined to:
\begin{equation} \label{eq8}
MDV = \left\{\varphi(X_{1}),\varphi(X_{2}),\cdots ,\varphi(X_{m})  \right\}
\end{equation}}

{We now explore distinct architectural patterns for evolving map systems as shown in Fig. \ref{fig4}. In the scenario previously observed in Fig. \ref{fig4}, rank pooling gets applied directly to the initial RGB video frames (see subsection \ref{sec_AHDI} ), arguably representing the "initial layer" in the system. This design is termed as the dynamic image system ( Fig. \ref{fig3}). On the other hand, the evolving map system in Fig. \ref{fig4} introduces rank pooling further up in the chain. This means implementing one or several feature processing layers on individual feature frames and then executing rank pooling on the resultant feature maps. Specifically, if we denote the feature maps formulated as :
\begin{equation} \label{eq66}
\hat{\rho}(F_1,F_2,\ldots,F_m; \varphi) = \sum^{m}_{\tau=1} \alpha_\tau\varphi(F_\tau)
\end{equation}
}
{The \( F_i \) represents the feature vector corresponding to the associated dynamic image \( \varphi(X_{1}) \), then rank pooling Eq \ref{eq66} is utilized to merge these maps into a unified evolving map. 
It parallels other layers like Max-pooling or Average-pooling, Its application is apt when there is a necessity to pool dynamic data over time. 
In addition to Rank-pooling, Max-pooling, Average-
pooling and LSTM \cite{graves2013speech}, we used other temporal pooling strategy such as :}

 {\noindent\textbf{Temporal Attention Pooling \cite{vaswani2017attention,xu2015show,yang2016hierarchical}:} 
    Uses attention weights for pooling across time. The weights determine the importance of each time step is features.
Given \( \left\{\varphi(X_{1}),\varphi(X_{2}),\cdots ,\varphi(X_{m})  \right\} \) a dynamic frames of same video \( V \), we represent the feature vector of each frame \( t \) as \( F_t \). After passing the dynamic frame through the base model (in this case, ResNet-50), we get these feature vectors 
\( (F_1,F_2,\ldots,F_m) \). For each feature vector \( F_t \), we pass it through a series of transformations to obtain its attention score \( a_t \):}
\begin{align}
a_t &= \text{Tanh}(W_1 F_t + b_1) \\
a_t &= W_2 a_t + b_2
\end{align}

{Where \( W_1 \), \( W_2 \) are weight matrices and \( b_1 \), \( b_2 \) are bias terms.
We then normalize these scores using a softmax function to get attention weights for each frame:
\begin{equation}
\alpha_t = \frac{e^{a_t}}{\sum_{i=1}^{T} e^{a_i}}
\end{equation}
Here, \( \alpha_t \) is the attention weight for frame \( t \). It signifies the importance of dynamic frame \( t \) relative to other dynamic frames in the video. Now, the feature representation for each frame is weighted using its corresponding attention weight:
\begin{equation}
F_t' = \alpha_t \times F_t
\end{equation}
Finally, the weighted feature vectors of all frames are summed up to produce a single feature vector \( F \) representing the entire video:}

\begin{equation}
F^* = \sum_{t=1}^{T} F_t'
\end{equation}
{The advantage of TAP is that it is trainable; during the training process, the attention weights are learned in a way.}
\section{Experiments}
In order to assess the efficacy of the proposed framework, we conducted exhaustive experiments on two benchmark datasets that are publicly accessible: the UNBC-McMaster Shoulder Pain Expression Archive \cite{UNBC-McMaster_Shoulder_Pain_Expression_Archive} and the BioVid Heat Pain database \cite{alghowinem2013ebiovid}. Our assessment begins with an analysis of the discrepancies between facial expressions in posed and genuine pain tasks, followed by pain level estimations. We further delve into a detailed evaluation of the performance of our selected model alongside the video summarizing methodology. We conclude the section by juxtaposing our findings with current state-of-the-art results.
\subsection{Experimental Data}
\noindent\textbf{UNBC-McMaster Archive:} 
The UNBC-McMaster Shoulder Pain Expression Archive \cite{UNBC-McMaster_Shoulder_Pain_Expression_Archive} stands as a pivotal database for pain expression research. Comprising videos that capture individuals' spontaneous facial reactions during various motion tests on their limbs, it offers invaluable insights. The archive provides FACS annotations for each frame in its video sequences, culminating in 16 unique pain intensity grades derived from action units. Our experiments predominantly revolve around the archive is active test set, incorporating 200 videos from 25 subjects, summing up to 48,398 frames.

\noindent\textbf{BioVid Heat Pain Database:}
The BioVid Heat Pain Database, referenced in Alghowinem {\it et al.} \cite{alghowinem2013ebiovid}, comprises 90 participants across three age groups, with bio-physiological metrics like SCL, ECG, EMG, and EEG accompanying visual recordings of varied pain intensities. We utilized data from Sections A and D: the former holds 8,700 labeled videos based on pain stimulus intensity from 87 participants,
\begin{figure}[ht]
    \centering
   \includegraphics[width=0.7\textwidth]{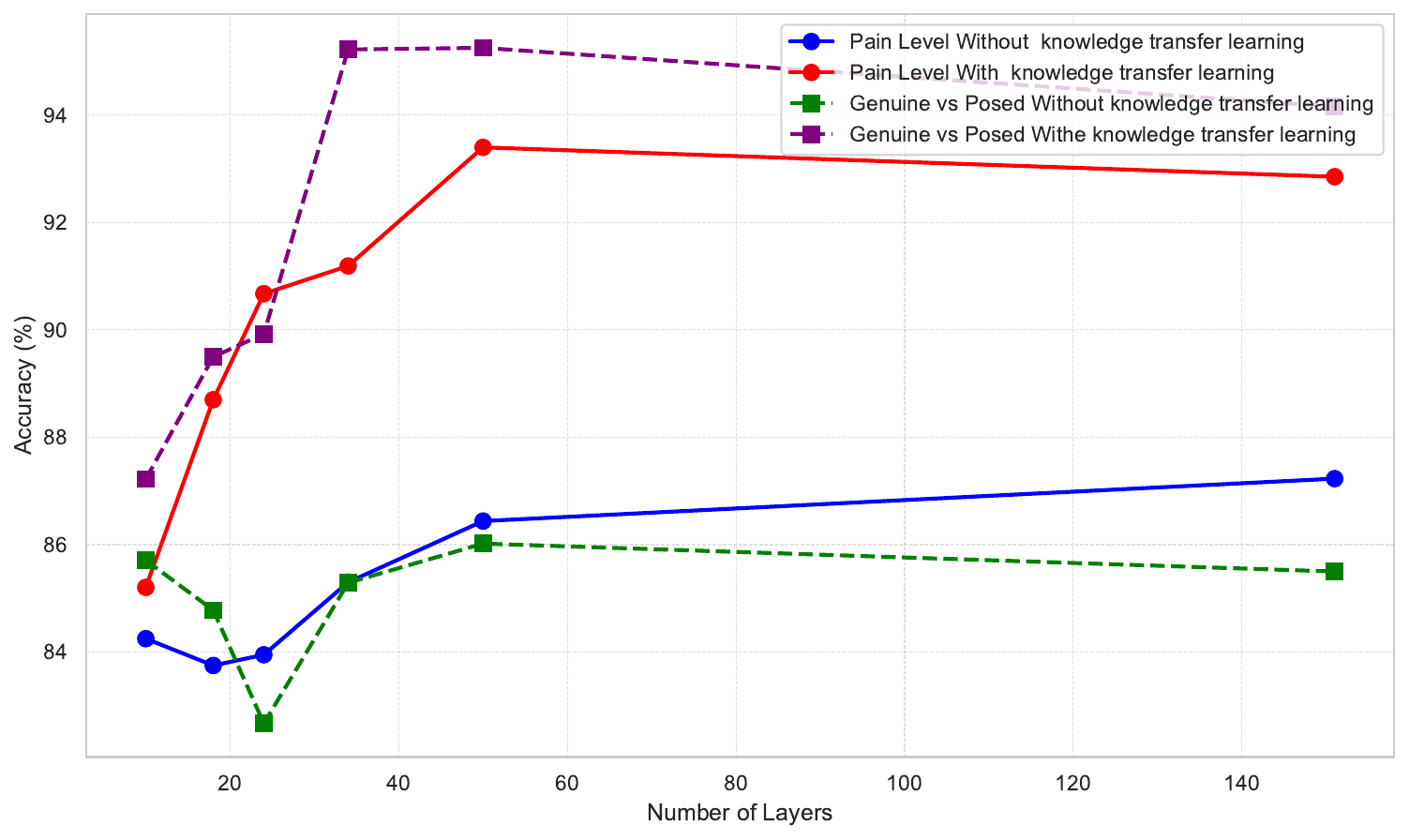}
    \caption{{Comparison of Accuracy with Different Number of Layers}}
    \label{accuracy_plot}
\end{figure}
and the latter offers 630 videos of posed facial expressions from 90 individuals. Unlike the UNBC-McMaster database, BioVid uses objective labeling from the heat-pain device is temperature and accounts for individual variability. Despite standardization, the 5.5-second videos sometimes contain neutral frames.  Evaluations on BioVid should emphasize the selection of expressive moments over these neutral states. Within this dataset, we focused on the binary classification between Neutral (BL1) and the most intense pain (PA4) states, encompassing 1740 videos for each category.

\subsection{Analysis of AHDI representations}
\textbf{Network Architecture Selection:}
In our quest to select the most suitable deep neural network architecture for pain detection in sync with AHDI representations, the ResNet series emerged as a promising candidate. Amongst them, ResNet-50 stood out, delivering exceptional performance as visualized in Fig. \ref{accuracy_plot} and quantitatively validated by Table \ref{ttab_2}. 

{Our investigations, graphically represented in Fig. \ref{accuracy_plot}, spanned various residual layer depths. The baseline performance, without knowledge transfer, was somewhat restricted, reaching an accuracy ceiling of 87\%. However, the integration of knowledge transfer brought about a remarkable surge, achieving 93.40\% accuracy with a 50-layer configuration for pain intensity discrimination.}

Furthermore, Table \ref{ttab_2} offers a comparative lens to evaluate our proposed approach against other prevalent models. A significant takeaway from this assessment is the clear advantage that AHDI transformation offers when paired with ResNet-50. The synergy between the two led to impressive MAE and MSE scores of \textbf{0.19} and \textbf{0.27} respectively, outperforming alternative architectures we considered.
\begin{table}[ht]
    \centering
    \begin{minipage}[t]{0.48\linewidth} 
        \centering
        \caption{Performance comparison of the proposed method and the generic models approaches using SDI/AHDI transformation}
        \begin{tabular}{l|l|l|l}
        \hline
         & Method & MAE & MSE \\ \hline
         \multirow{7}{*}{No SDI} & Inception \cite{szegedy2015going} & 0.73 &  1.72\\ 
         & xception \cite{chollet2017xception} &0.68&1.71  \\ 
         & EfficientNetV3 \cite{tan2021efficientnetv3} &0.66 &1.63  \\ 
         & Vit \cite{dosovitskiy2020vit} &  0.72&  1.69\\ 
         & Swin-Vit \cite{liu2021swinvit} &0.90  &1.81  \\ 
         & ResNet50 \cite{he2016resnet}&0.57  & 1.60 \\ 
         \hline\hline
        \multirow{7}{*}{SDI} & Inception \cite{szegedy2015going} & 0.40 & 1.21 \\ 
         & xception \cite{chollet2017xception} & 0.30 & 0.33 \\ 
         & EfficientNetV3 \cite{tan2021efficientnetv3} & 0.29 & 0.33 \\ 
         & Vit \cite{dosovitskiy2020vit} & 0.32 & 0.84 \\ 
         & Swin-Vit \cite{liu2021swinvit} & 0.34 & 0.98 \\ 
        & ResNet50\cite{he2016resnet}& 0.31 & 0.45 \\ 
         \hline
         \multirow{1}{*}{AHDI} 
        & \textbf{ResNet50} \cite{he2016resnet}& 0.19 & 0.27 \\  \hline
        \end{tabular}
        \label{ttab_2}
    \end{minipage}
    \hfill 
    \begin{minipage}[t]{0.48\linewidth} 
        \centering
        \caption{Performance comparison of various single image video representation methods in terms of Accuracy (\%)}
        \begin{tabular}{lc}
        \hline
        Descriptor & Acc (\%) \\ \hline
        Mean Image &79.90  \\ \hline
        OF – Optical Flow & 81.31 \\ \hline
        {AHDI} &  93.40\\ \hline
        \end{tabular}
        \label{tab2}
    \end{minipage}
\end{table}

\textbf{Data Preprocessing and Representation Extraction:}
For an exhaustive and comprehensive analysis, Through a meticulous decomposition of each video into its fundamental frames, we retained inherent frame rates. We expanded our representation by incorporating optical flow, guided by the method in \cite{farneback2003two}. AHDI is proficiency was critically assessed against established methods like Mean, Optical Flow. All ranking processes were preemptively executed offline, thus enhancing computational efficiency. When these dynamic images were evaluated using ResNet50, as depicted in Table \ref{tab2}, AHDI emerged superior with an accuracy of 93.40\%. This underscores rank pooling is potential superiority over traditional methods. Such findings suggest a paradigmatic shift towards advanced techniques like AHDI for future video analysis, especially given the burgeoning complexity of video data.

The AHDI method, detailed in Section \ref{sec_AHDI}, employs an innovative overlapping strategy synergized with adaptive segmentation for precise video sequence decomposition. This combination emphasizes volumetric disparities between successive video segments, allowing AHDI to achieve superior spatiotemporal pooling. Such a design enables AHDI to represent not just visual nuances but also the dynamism inherent in video sequences, condensing a video is entirety into one image while maintaining channel consistency. The method is success is contingent on the judicious adjustment of two parameters: segment temporal length and patch spatial dimensions.

Table \ref{tab_4} underscores AHDI is robustness. The table evaluates the method is prowess in pain intensity classification and distinguishing genuine from posed pain expressions over various segment lengths. A notable observation is the 8\% accuracy boost when segment length is fixed at 25 frames. Transitioning to dynamic segmentation length further amplified performance: witnessing a 6\% rise in pain intensity accuracy and a remarkable 10\% improvement in discerning genuine pain expressions.

\begin{table}[ht]
    \centering
    \begin{minipage}[b]{0.45\linewidth} 
        \centering
        \caption{Performance comparison with different Spatial patch size.}
        \begin{tabular}{lcc}\hline
            & Pain Level & Genuine versus Posed Pain \\\hline
            [3x3]  & 84.56 & 88.20 \\\hline
            [5x5]  & 87.02 & 91.07 \\\hline
            [7x7]  & 91.44 & 90.32 \\\hline
            [9x9]  & 93.40 & 94.03 \\\hline
            [11x11] & 91.76 & 92.51 \\\hline
            [15x15] & 90.48 & 88.30 \\\hline
            [17x17] & 88.60 & 87.43 \\\hline                        
        \end{tabular}
        \label{tab_3}
    \end{minipage}
    \hfill 
    \begin{minipage}[b]{0.45\linewidth} 
        \centering
        \caption{Performance comparison of the proposed method versus different lengths of Temporal segments.}
        \begin{tabular}{lcc}\hline
            & Pain Level & Genuine versus Posed Pain \\\hline
            5 & 83.36 & 79.20 \\\hline
            10 & 80.88 & 80.21 \\\hline
            15 & 84.41 & 82.50 \\\hline
            20 & 87.24 & 84.30 \\\hline
            25 & 87.15 & 84.66 \\\hline
            30 & 85.60 & 84.56 \\\hline
            40 & 82.90 & 81.98 \\\hline
            AHDI & 93.40 & 94.03 \\\hline 
        \end{tabular}
        \label{tab_4}
    \end{minipage}
\end{table}

As the segment length extended beyond 30 frames, we observed a decrement in performance. This suggests that protracted segments might unintentionally capture extraneous spatiotemporal details, diluting the ability to detect subtle variances within the video. Notwithstanding, the pinnacle of precision was attained using AHDI is inherent adaptive segmentation technique.

The patch spatial dimensions in the AHDI methodology critically influence its efficacy, as illustrated in Table \ref{tab_3}. Smaller window sizes restrict the receptive field, thereby potentially overlooking crucial spatial relationships within the video. On the other end of the spectrum, overly expansive patches can reduce accuracy due to information over-aggregation, resulting in a representation that might be overly generalized. Empirical analysis identifies a 9x9 patch size as the optimal dimension for AHDI is performance.

Perusing Table \ref{tab_3} brings to light that the disparities in classification accuracies are more nuanced as compared to the stark variations in adversarial accuracies. This alludes to the resilience of the network is classification mechanism, even in the face of marginal perturbations in input data. However, the precise detection of genuine pain expressions is contingent on subtle facial nuances. Hence, engendering a meticulous representation of facial appearances and dynamics is paramount for the efficacious discernment between authentic and contrived facial expressions.

 {\textbf{Multiple Vs Single Dynamic Visualizations:}
In our initial approach, a single dynamic image was used to represent a video, as depicted in Fig. \ref{fig2}-\ref{fig3}. Yet, our investigation indicates a potential enhancement by dividing a video into multiple sub-sequences and representing each with a dynamic image. The rationale behind this was to discern its impact on classification accuracy. However, while MDV is less accurate than SDV,
it is also slower due to the fact that MDV extract features from a
number of dynamic images proportional to the video length, so SDV can be preferable when speed is paramount. Delving deeper into the temporal pooling layers labeled "temp-pool", we found that Temporal Attention Pooling stands out, achieving the pinnacle of video classification accuracy at 84.41\% as described in Tabel \ref{tab_77}. This superior performance can be ascribed to its inherent ability to remain invariant to temporal positioning and ensure consistent action alignment throughout various videos. In contrast, other methods like Average Pooling, Max Pooling, Rank Pooling, and LSTM did not display the same level of efficiency.}
\begin{table}[ht]
    \centering
    \begin{minipage}[t]{0.48\linewidth} 
        \centering
        \caption{Performance comparison of single vs multiple dynamic image networks in terms of Accuracy (\%)}
        \begin{tabular}{l|l}
        \hline
        Method & Acc  \\\hline
        Average Pooling & 40.10  \\
        Max Pooling & 44.31 \\
        Rank Pooling & 70.11 \\
        LSTM & 79.07 \\
        Temporal Attention Pooling & 84.41 \\  
        \textbf{AHDI} & 89.76\\  \hline
        \end{tabular}
        \label{tab_77}
    \end{minipage}
    \hfill 
    \begin{minipage}[t]{0.48\linewidth} 
        \centering
        \caption{Performance comparison of the proposed method and the state-of-the-art approaches}
        \begin{tabular}{l|l|l}
        \hline
        Method & MAE & MSE \\\hline
RCNN  \cite{zhou2016recurrent}&- &1.54\\  
LSTM  \cite{7849133}  &-&0.74 \\
 C3D \cite{wang2018pain}& -& 0.88\\  
 WSDA-OR  \cite{praveen2020deep}&0.53  &  -  \\  
 SSD \cite{Tavakolian2020}  &  -& 0.92	  \\  
 PET \cite{xu2021deep} & - & 0.40  \\  
\textbf{{AHDI+ResNet50}} & 0.19& 0.27 \\  \hline
        \end{tabular}
        \label{tab_7}
    \end{minipage}
\end{table}

\subsection{Comparison against the state-of-the-art}
{Within the rapidly advancing milieu of pain estimation methodologies, periodic evaluations against stalwarts are paramount. Thus, our innovative approach was pitted against the notable benchmarks, namely, C3D \cite{wang2018pain}, WSDA-OR \cite{praveen2020deep}, SSD \cite{Tavakolian2020}, and PET \cite{xu2021deep}.}
Our empirical investigations were bifurcated:
\begin{enumerate}
    \item {Deploying our framework exclusive of AHDI, to glean insights into its inherent capabilities.}
    \item {Integration of AHDI, to discern its augmentative impact, especially in the nuanced realm of facial dynamics – a sine qua non for differentiating authentic from simulated pain expressions.}
\end{enumerate}

{Table \ref{tab_7} offers an analytical distillation, juxtaposing our method with its contemporaries, utilizing the metrics of Mean Absolute Error (MAE) and Mean Squared Error (MSE). Evidently, our AHDI+ResNet50 synergy manifests superlative metrics, with an MAE and MSE of 0.19 and 0.27, respectively. This dominance is accentuated when compared to the esteemed PET \cite{xu2021deep} which posts an MSE of 0.40.
Aiming to position our AHDI+ResNet50 technique amidst established methodologies, we further extended our evaluation on the widely recognized Biovid dataset. This dataset, renowned for its heterogeneity and rich annotations, offers a rigorous testing ground for pain estimation techniques, making its employment for benchmarking a judicious choice.}

{Table \ref{tab_8} presents a comparative scrutiny of our model is performance against various leading-edge methods in terms of accuracy (\%). The methods incorporated for this comparison encompass a spectrum of techniques from texture-based features like LBP \cite{yang2016pain}, LPQ \cite{yang2016pain}, and BSIF \cite{yang2016pain}, to more complex methodologies like BORMIR \cite{zhang2018bilateral} and the recent works of Lopez {\it et al.} \cite{lopez2017multi} and Ayral {\it et al.} \cite{ayral2021temporal}. A salient observation from this table is the superior accuracy of our AHDI+ResNet50 methodology, registering an impressive 89.76\%. This score not only underscores its efficacy but also its supremacy in handling the diverse challenges posed by the Biovid dataset. The technique outpaces its closest competitor by over 5\%, attesting to its advanced capability to discern genuine pain expressions amidst a myriad of facial dynamics.}

{A perusal of Table \ref{tab_9} highlights the seminal role of AHDI. With its incorporation, the framework is accuracy swells to an illustrious 94.03\%, notably in the discernment between genuine and simulated pain expressions. For context, the standalone ResNet50 manages a modest 35.24\%. The divergence in these outcomes underscores AHDI is pivotal contribution to model enhancement, a sentiment further echoed by AHDI+ResNet50 is supremacy over other pioneering methods.}

\begin{table}[ht]
    \centering
    \begin{minipage}[t]{0.48\linewidth} 
        \centering
        \caption{Performance comparison of the proposed method and the state-of-the-art approaches on Biovid dataset}
        \begin{tabular}{l|l}
        \hline
        Method & Acc  \\\hline
LBP \cite{yang2016pain} & 63.72 \\
LPQ \cite{yang2016pain} & 63.19 \\
BSIF \cite{yang2016pain} & 65.17 \\
FAD set \cite{werner2016automatic} & 72.40 \\
BORMIR \cite{zhang2018bilateral} & 72.85 \\
Lopez {\it et al.} \cite{lopez2017multi} &82.75\\
Ayral {\it et al.} \cite{ayral2021temporal} & 82.67 \\
Ayral {\it et al.} \cite{ayral2021temporal} & 84.39 \\
\textbf{AHDI+ResNet50} & \textbf{89.76} \\ \hline
        \end{tabular}
        \label{tab_8}
    \end{minipage}
    \hfill 
    \begin{minipage}[t]{0.48\linewidth} 
        \centering
        \caption{Performance comparison of the proposed method and the state-of-the-art approaches (Genuine versus Posed Pain)}
        \begin{tabular}{l|c}
        \hline
        Method & Acc \\\hline
{ResNet50} & 35.24\\ 
 R-GAN WSP \cite{tavakolian2019learning}& 85.05 \\  
 \hline \hline
\textbf{AHDI+ResNet50} &\textbf{94.03}\\\hline
        \end{tabular}
        \label{tab_9}
    \end{minipage}
\end{table}

\subsection{Conclusion}
In this study, we unveiled the Adaptive Hierarchical spatio-temporal Dynamic Image (AHDI) method, a transformative approach that condenses spatiotemporal facets of facial videos into a single RGB image. Leveraging a residual network, we crafted facial representations tailored for dual purposes: quantifying pain intensity and distinguishing genuine from posed pain expressions. Testing on two prominent pain datasets yielded promising results, rivaling contemporary benchmarks. In the expansive domain of affective computing, the automatic emotion recognition from facial expressions remains a pivotal topic of academic exploration. The efficacy of these cutting-edge methods ties closely to the specificity of their application and the depth of available data. Advancements in autonomous pain assessment have been anchored around three core principles: (1) distinguishing features from single and combined computational models, (2) optimizing the use of limited and ordinal data, and (3) developing tailored models emphasizing temporal patterns. Despite progress, a gap remains between academic innovations and real-world clinical applications. 
Looking forward, two promising directions emerge: (i) leveraging pre-trained models and adapting them to pain recognition tasks due to ongoing data challenges, and (ii) fostering collaboration between technologists, clinicians, and psychologists to develop more context-sensitive systems.

\bibliographystyle{unsrt}  

\bibliography{main}

\end{document}